\documentclass[10pt,twocolumn,letterpaper]{article}

\usepackage{cvpr}              % To produce the CAMERA-READY version
\usepackage{pifont}               % Required for \ding command
\newcommand{\cmark}{\ding{51}}   % Checkmark
\newcommand{\xmark}{\ding{55}}   % Crossmark

\usepackage{colortbl}             % Required for column colors in tables
\usepackage{makecell}             % For line breaks in table cells

\usepackage{algorithm}            % Required for algorithm environment
\usepackage{algpseudocode}       % Required for algorithmic environment
\usepackage[dvipsnames]{xcolor}

\definecolor{cvprblue}{rgb}{0.21,0.49,0.74}
\usepackage[pagebackref,breaklinks,colorlinks,allcolors=cvprblue]{hyperref}

\def\paperID{38337} % *** Enter the Paper ID here
\def\confName{CVPR}
\def\confYear{2026}

\title{PDD: Manifold-Prior Diverse Distillation for Medical Anomaly Detection}

\author{Xijun Lu$^{1}$, Hongying Liu$^{1*}$, Fanhua Shang$^{2}$, Yanming Hui$^{2}$, Liang Wan$^{1,2}$\\
$^{1}$Medical School, Tianjin University, Tianjin, China\\
$^{2}$College of Intelligence and Computing, Tianjin University, Tianjin, China\\
{\tt\small\{xjunlu, hyliu2009, fhshang, ymhui, lwan\}@tju.com}
}
\begin{document}
\maketitle
\begin{abstract}
Medical image anomaly detection faces unique challenges due to subtle, heterogeneous anomalies embedded in complex anatomical structures. Through systematic Grad-CAM analysis, we reveal that discriminative activation maps fail on medical data, unlike their success on industrial datasets, motivating the need for manifold-level modeling. We propose \textbf{PDD} (Manifold-Prior Diverse Distillation), a framework that unifies dual-teacher priors into a shared high-dimensional manifold and distills this knowledge into dual students with complementary behaviors. Specifically, frozen VMamba-Tiny and wide-ResNet50 encoders provide global contextual and local structural priors, respectively. Their features are unified through a \textbf{Manifold Matching and Unification (MMU)} module, while an \textbf{Inter-Level Feature Adaption (InA)} module enriches intermediate representations. The unified manifold is distilled into two students: one performs layer-wise distillation via \textbf{InA} for local consistency, while the other receives skip-projected representations through a \textbf{Manifold Prior Affine (MPA)} module to capture cross-layer dependencies. A diversity loss prevents representation collapse while maintaining detection sensitivity. Extensive experiments on multiple medical datasets demonstrate that \textbf{PDD} significantly outperforms existing state-of-the-art methods, achieving improvements of up to 11.8\%, 5.1\%, and 8.5\% in AUROC on HeadCT, BrainMRI, and ZhangLab datasets, respectively, and 3.4\% in F1 max on the Uni-Medical dataset, establishing new state-of-the-art performance in medical image anomaly detection. The implementation will be released at {\color{WildStrawberry}\url{https://github.com/OxygenLu/PDD}}.
\end{abstract}    
\begin{figure}[t]
  \centering
  \includegraphics[width=0.95\linewidth]{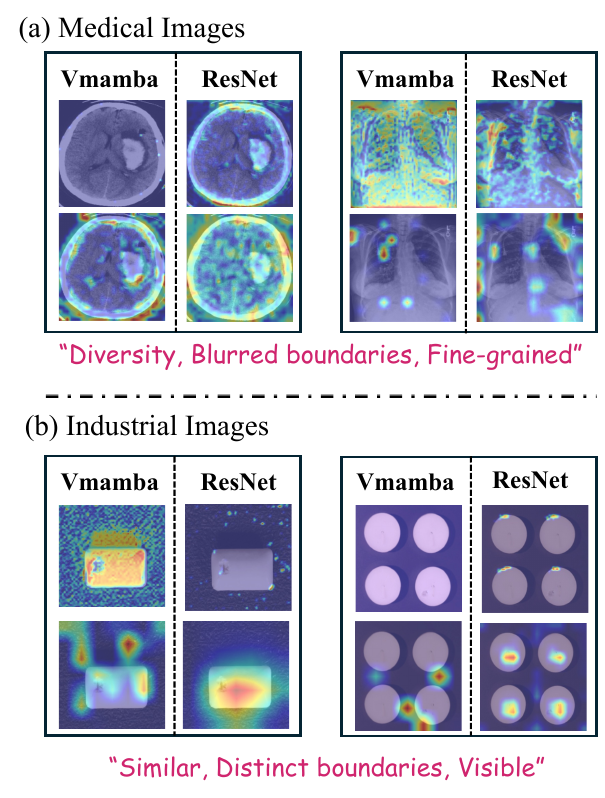}
   \caption{\textbf{Grad-CAM visualization of frozen Vmamba and ResNet across medical and industrial images.} Within each group, feature maps progress from low-dimensional to high-dimensional feature representations (top to bottom). At the same feature dimension, the dispersed and aggregated activation patterns of the two backbones form complementary prior information.}
   \label{fig:mean}
\end{figure}
% \vspace{-0.em}

\section{Introduction}
\label{sec:intro}

Medical image anomaly detection has attracted increasing attention as a crucial component for early disease screening and computer-aided diagnosis~\cite{ZUO,li2026causal,MedIAnomaly, l2023unsupervised}. Unlike conventional anomaly detection on natural or industrial images, medical anomalies often exhibit subtle, heterogeneous, and weakly contrasted patterns embedded in highly complex anatomical structures ~\cite{iqbal2023unsupervised, lagogiannis2023unsupervised}. These characteristics make unsupervised anomaly detection (UAD) particularly challenging, since models must learn a compact and stable manifold of normal anatomy solely from healthy samples, without access to annotated abnormal regions~\cite{bercea2025evaluating}.

Recent teacher--student frameworks have shown promising performance in anomaly detection, especially in industrial inspection tasks~\cite{bergmann2020uninformed,rd4ad, rrd4ad, gu2023remembering}. However, as shown in Fig.~\ref{fig:mean}, Grad-CAM visualizations reveal a striking contrast: on industrial datasets such as MVTec, Grad-CAM produces clean and highly localized heatmaps, while on medical datasets (e.g., BrainMRI, HeadCT), the heatmaps become diffuse, noisy, and anatomically inconsistent~\cite{l2023unsupervised,MedIAnomaly}. This observation reveals a fundamental difference: industrial defects are typically texture-driven and spatially localized, while medical anomalies are structural deviations distributed across anatomical hierarchies with subtle, context-dependent boundaries~\cite{huang2024adapting, gu2023remembering, guo2023recontrast,iqbal2023unsupervised}. As a result, single-stream feature extractors are insufficient to learn a complete and anatomically coherent normal manifold.

These findings motivate the need for a principled approach that can integrate multiple complementary priors to learn a robust and unified normal manifold for medical images~\cite{squid,mamba,simsid,StegOT}. CNNs excel at capturing fine-grained local textures~\cite{h2former,lin2024ratlip,liu2025consistency,chen2024transunet, mcintosh2023inter}, whereas sequence models such as Mamba capture long-range dependencies and global structural patterns~\cite{mambaad,liu2024fedbcgd,wu2025cmanet,lin2025mtmamba,sun2025spatial,huang2025enhancing,liu2024swin}. However, directly fusing their features does not guarantee alignment between their heterogeneous manifolds, nor does it ensure that downstream student networks preserve representation diversity---an important property for reliably detecting subtle abnormal deviations~\cite{destseg,mambaad,uniad}.

To address these challenges, we propose \textbf{PDD} (Manifold-Prior Diverse Distillation), a framework that unifies dual-teacher priors into a shared high-dimensional manifold and distills this unified knowledge into dual students with complementary learning behaviors. Specifically, a frozen VMamba-Tiny encoder and a frozen wide-ResNet50 encoder provide global contextual and local structural priors, respectively. Their multi-scale features are aligned through the \textbf{Manifold Matching and Unification (MMU)} module, while the \textbf{Inter-Level Feature Adaption (InA)} module fuses intermediate features from both teachers. This unified manifold is then distilled into two structurally identical but functionally diverse students: Student 1 performs layer-wise distillation from the fused teacher features via \textbf{InA}, focusing on local consistency, while Student 2 receives skip-projected latent representations from the unified manifold through the \textbf{Manifold Prior Affine (MPA)} module, enabling it to capture cross-layer contextual dependencies. A diversity loss ensures agreement on normal structures while allowing diversity in detecting potential abnormalities.

Our contributions can be summarized as follows:
\begin{itemize}
    \item We propose a novel dual-teacher architecture that leverages complementary representations from heterogeneous backbones (VMamba-Tiny for global contextual priors and wide-ResNet50 for local structural priors), addressing the limitations of single-stream feature extractors in medical anomaly detection.
    \item We introduce a dual-teacher manifold unification module (\textbf{MMU}) that integrates global contextual priors from VMamba-Tiny and local structural priors from wide-ResNet50 into a cohesive high-dimensional anatomical manifold.
    \item We propose a dual-student diverse distillation strategy that enhances representation stability by combining local distillation (via \textbf{InA}), cross-layer manifold projection (via \textbf{MPA}), and dual-student consistency regularization, achieving significant improvements over existing methods on multiple medical datasets.
\end{itemize}

% Please follow the steps outlined below when submitting your manuscript to the IEEE Computer Society Press.
% This style guide now has several important modifications (for example, you are no longer warned against the use of sticky tape to attach your artwork to the paper), so all authors should read this new version.
% 中文
% 监督视觉异常检测 (UAD)，通常可以看作求
% 解偏离正常图像模式建模的异常模式。通常异常区
% 域具有罕见性，不同数据域多样化的特点，如图像
% 目标的缺陷，病变，位置错误等异常。异常检测在工

%-------------------------------------------------------------------------

\section{Related Work}
\subsection{Visual anomaly detection}
Visual anomaly detection can be framed as identifying deviations from learned normal sample patterns. In recent years, numerous effective, domain-specific anomaly detection models have emerged in both the medical and industrial fields. CutPaste~\cite{cutpaste}, a method developed for industrial image analysis, introduced an auxiliary task (image cropping and rotation) to simulate anomalies, pioneering the application of self-supervised learning in anomaly detection. For instance, f-AnoGAN~\cite{f-AnoGAN}, which is based on adversarial generative networks, identifies anomalies using discriminators operating on feature vectors in latent space; however, its iterative optimization process is inefficient, and the generator is prone to mode collapse.

In addition, recent efforts have focused on developing unified models for anomaly detection. For example, UniAD~\cite{uniad} employs a patch-based reconstruction method optimized for industrial applications; however, its performance degrades in other domains and requires a large number of normal samples for training. UniVAD~\cite{univad} utilizes pre-trained Grounding DINO~\cite{dino} and SAM~\cite{SAM}, achieving training-free anomaly detection and demonstrating satisfactory performance on cross-domain datasets.

In the field of medical image analysis, SQUID~\cite{squid} and SimSID~\cite{simsid}, proposed by Tiange Xiang et al., represent successive works. These models incorporate a memory storage module and transformer positional encoding, proving effective for medical images with fixed structures, such as X-rays, but ineffective at decoupling anatomical structural information during dimensionality reduction processes like CT (3D to 2D).

\begin{figure*}[t]
    \centering
    \includegraphics[width=0.95\linewidth]{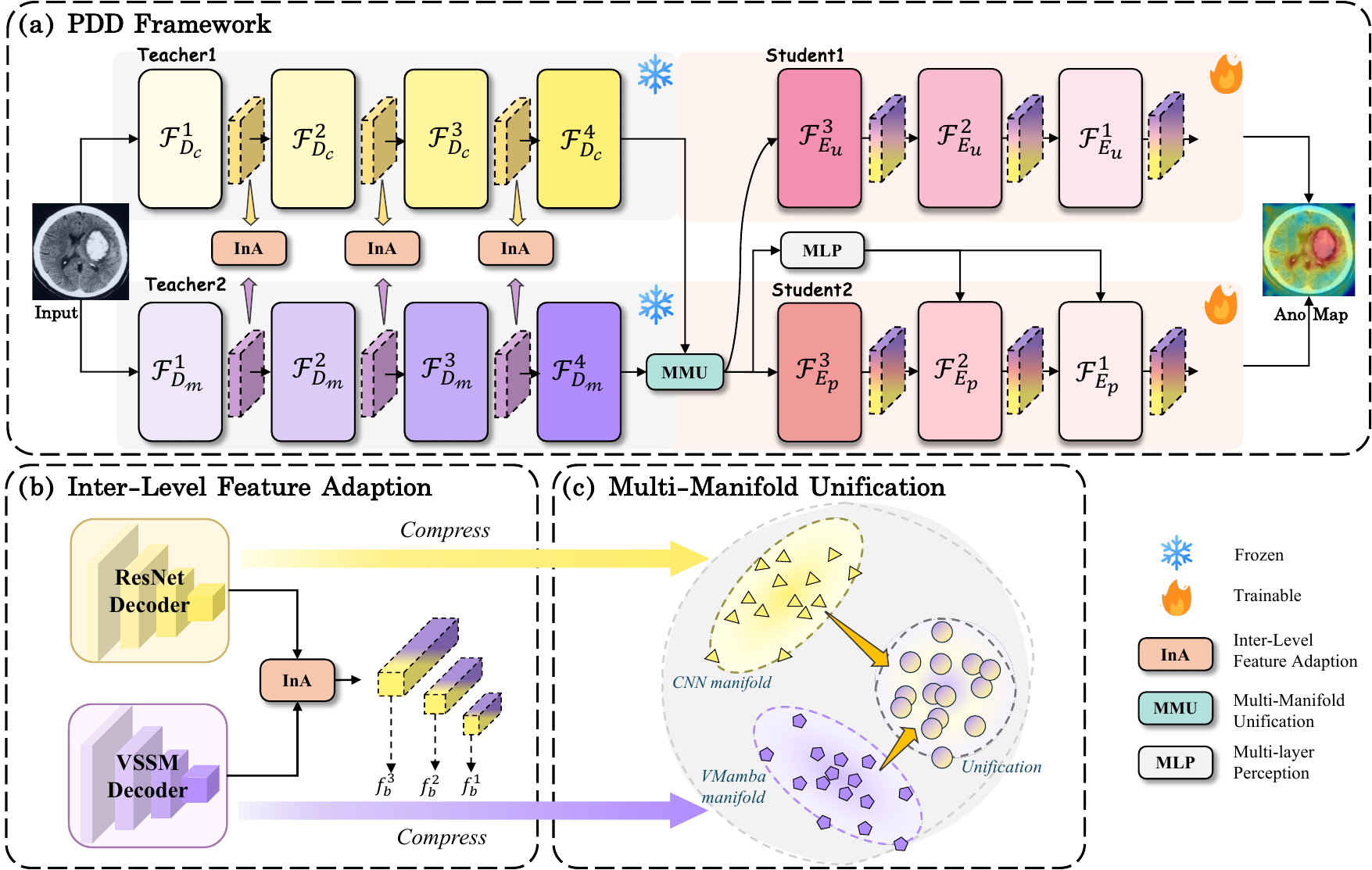}
     \caption{\textbf{Overview of the proposed PDD framework. }The framework employs a dual-teacher and dual-student architecture. The teachers consist of frozen VMamba-Tiny and frozen wide-ResNet50 encoders, whose intermediate features are fused via the InA module (shown in (b)) to obtain $f_{b}^i$. The two teacher encoders compress input images into distinct high-dimensional manifold spaces, which are then aligned through the MMU module. The aligned features are fed into two student networks: Student 1 distills features from InA via $\mathcal{F}_{E_{u}}^i$, while Student 2 incorporates multi-scale manifold space features from the unified manifold through MLP-based skip connections to $\mathcal{F}_{E_{p}}^i$, distilling both prior and InA features. This enables diverse reconstruction of normal samples and effective separation of anomalies.}
     \label{fig:onecol}
  \end{figure*}
% \vspace{-0em}

\subsection{Methods Based on Knowledge Distillation}

In recent years, knowledge distillation based anomaly detection models have emerged as a promising method for balancing computational efficiency with model performance. The traditional knowledge distillation (KD) paradigm involves transferring knowledge from a teacher network to a student network for the identification and localization of anomalous samples. Typically, teacher and student networks adopt similar or identical architectures, resulting in insufficient representation diversity when processing out-of-distribution samples. RD4AD~\cite{rd4ad} pioneered a new paradigm termed reverse knowledge distillation, in which the student network learns directly from the teacher's high-level features, progressively capturing multi-scale information across the teacher's layers to enable the efficient separation of normal from anomalous samples. Inspired by RD4AD~\cite{rd4ad}, SKip-TS~\cite{skip-ts} employs skip connections and multi-scale feature sampling between the teacher and student networks, thereby mitigating the challenge of normal pattern reconstruction prevalent in medical datasets—a limitation identified in RD4AD. More recently, CFRG~\cite{yang20253cad} has been introduced, leveraging heterogeneous teacher-student networks to extract multi-scale features, utilizing recovery networks to reconstruct normal patterns from anomalous features, and enabling fine-grained anomaly localization through segmentation networks. Our method introduces a novel reverse knowledge distillation framework, aiming to enhance the consistency of the student networks to achieve more effective separation of normal from anomalous patterns.

Some studies have found that pre-trained networks on large datasets such as ImageNet-1k can serve as backbones for anomaly detection. Models like SAM, Mamba, and DINO leverage their ability to extract features across images, enabling direct anomaly detection on samples. However, this computational efficiency comes at the cost of reduced feature extraction capability. UniVAD, as previously mentioned, requires explicit mask labeling for its target objects. Consequently, its performance on datasets lacking such annotations or characterized by fine-grained normal regions—such as BrainMRI~\cite{brainmri} and CheXray~\cite{wang2017chestx}—exhibits a significant gap compared to domain-specific models.

However, existing methods typically employ single-teacher architectures and focus on homogeneous feature spaces, limiting their ability to capture diverse representations of normal patterns. Our work addresses these limitations by introducing a dual-teacher dual-student framework that leverages heterogeneous backbones and unified manifold spaces to enable more diverse and robust normal pattern learning.

\section{Methodology}
\subsection{Manifold-Unified Reverse Distillation}
As shown in Fig.~\ref{fig:onecol}, we propose \textbf{PDD}, a manifold-unified reverse distillation framework that employs dual-teacher and dual-student architecture to leverage complementary representations from heterogeneous backbones. The teachers consist of frozen VMamba-Tiny and frozen wide-ResNet50 encoders, which process input images in parallel and extract multi-scale intermediate features. These features are fused via the \textbf{InA} module (Fig.~\ref{fig:onecol}-(b)) to obtain enriched fused features $f_{b}^i$. The two teacher encoders compress input images into distinct high-dimensional manifold spaces: VMamba-Tiny learns sequential state-space representations while wide-ResNet50 captures spatial convolutional features. The \textbf{MMU} module (Fig.~\ref{fig:onecol}-(c)) performs geometric alignment between these heterogeneous manifolds, mapping semantically similar features to a unified manifold space. The aligned features are fed into two student networks: Student 1 distills features from \textbf{InA} via $\mathcal{F}_{E_{u}}^i$ for unified representations, while Student 2 incorporates multi-scale manifold features through MLP-based skip connections to $\mathcal{F}_{E_{p}}^i$, distilling both prior knowledge from the unified manifold and enriched features from \textbf{InA}. This dual-student design enables diverse reconstruction of normal samples while maintaining sensitivity to anomalies.

\subsection{Intra-Backbone Feature Fusion Strategy}
%我们采用两个具备大规模图像数据预训练权重的异构骨干网络Wide-ResNet50(ResNet50)和vanilla-vmamba-tiny（Vmamba-Tiny）作为教师模型，即teacher1和teacher2。这两个backbone在ImgaeNet-1k上都取得了令人兴奋的效果，它们的预训练权重具有很强的图像特征捕获能力。

We employ two heterogeneous backbone networks with large-scale pre-trained weights as teacher models: wide-ResNet50 and vanilla-VMamba-Tiny (VMamba-Tiny), denoted as Teacher 1 and Teacher 2, respectively. Both backbones have demonstrated strong performance on ImageNet-1K, and their pre-trained weights exhibit powerful image feature extraction capabilities. As shown in Fig.~\ref{fig:mean}, Grad-CAM visualizations reveal distinct feature representations at shallow and deep layers for both ResNet and Mamba backbones. The complementary nature of these representations suggests that fusing shallow features from wide-ResNet50 and VMamba-Tiny can significantly enrich the prior knowledge at each network layer, thereby obtaining a more powerful feature extractor.
% 在图\ref{cam}中，Grad-CAM分别对ResNet和Mamba的backbone的浅层次结构
% 和深层次结构进行了热力图可视化。根据我们在前文的证明，可以很直观的知道
% wide-ResNet和Vmamba进行浅层特征融合，backbone层的特征融合，可以极大程度
% 的丰富每一层网络的先验知识，进而获得一个更加强大的特征提取器。
% \begin{figure}[t]
%   \centering
%   \includegraphics[width=0.95\linewidth]{figure/detail.pdf}
%    \caption{Detailed implementation of the InA, MMU and Manifold Prior Affine(MPA) module.}
%    \label{fig:detail}
% \end{figure}

To this end, we design a lightweight inter-backbone feature fusion adapter, termed InA (Inter-Level Feature Adaption). Formally, we define the input image as $x_{n} \in \mathbb{R}^{C \times H \times W}$, which is reshaped and fed into the two teacher encoders. The VMamba backbone, as shown in Fig.~\ref{fig:onecol}, consists of 4 blocks denoted as $\mathcal{F}_{D_{m}}^{i}$ for $i \in [1, 4]$. The frozen vanilla-VMamba-Tiny pre-trained weights output a feature map $f_{m}^{i} \in \mathbb{R}^{h \times w \times c}$ between each layer, where $h$, $w$, and $c$ represent the height, width, and number of channels, respectively. Similarly, the wide-ResNet50 branch consists of 4 blocks, denoted as $\mathcal{F}_{D_{c}}^{i}$ for $i \in [1, 4]$, which output features $f_{c}^{i} \in \mathbb{R}^{h' \times w' \times c'}$ between layers.

The shallow feature fusion process between the two backbones is performed as follows:
\begin{equation}
     S(f_{m}^{i}, f_{c}^{i}) = \left( \frac{h'}{h},\ \frac{w'}{w} \right), \quad
    \tilde{f}_{m}^{i} = \mathcal{U}(f_{m}^{i}, S)
\end{equation}
\begin{equation}
    f_{b}^{i} = \mathcal{F}_{\text{InA}}\left(f_{m}^{i}, f_{c}^{i}\right) = \tilde{f}_{m}^{i} + f_{c}^{i}
\end{equation}
where $S(\cdot)$ is a scaling function that computes the spatial scaling factors, $\mathcal{U}(\cdot)$ is an upsampling function that performs 2D bilinear interpolation to scale the Mamba feature map to match the convolutional feature map dimensions, and $\mathcal{F}_{\text{InA}}(\cdot)$ implements the feature fusion at shallow layers, producing the fused feature $f_{b}^{i}$.

\subsection{Manifold Space Unification of Heterogeneous Backbones}

The input image $x_n$ is processed by the two teacher encoders in parallel, and through their respective encoding processes, the image is compressed into high-dimensional manifold spaces, as illustrated in Fig.~\ref{fig:onecol}-(c). Due to their distinct architectural inductive biases, the two teacher encoders naturally embed input images into different high-dimensional manifold spaces. The VMamba-Tiny encoder, with its state-space modeling mechanism, learns representations in a sequential state-space manifold $\mathcal{M}_m \subset \mathbb{R}^{d_m}$, while the wide-ResNet50 encoder, with its convolutional structure, captures features in a spatial convolutional manifold $\mathcal{M}_c \subset \mathbb{R}^{d_c}$, where $d_m$ and $d_c$ denote the dimensions of the respective manifolds. These heterogeneous manifolds encode complementary semantic information but exist in different geometric spaces, making direct feature interaction challenging.

The deep features at the tail of the backbones contain rich high-level semantic information that is crucial for understanding the overall image semantics. To fully leverage the high-level semantic features from both teacher networks and unify the heterogeneous manifolds into a common space, we propose the \textbf{M}anifold \textbf{M}atching and \textbf{U}nification (MMU) module. The MMU module is responsible for fusing the high-level semantic features from both teacher models, thereby strengthening the modeling capability for overall image semantic consistency.

Formally, let $f_{m}^{i} \in \mathbb{R}^{h \times w \times c}$ and $f_{c}^{i} \in \mathbb{R}^{h' \times w' \times c'}$ denote the tail features extracted from the VMamba and ResNet encoders at layer $i$, respectively. The MMU module first adapts the Mamba features through a channel-wise adaptation pathway. The MMU consists of a $1 \times 1$ dilated convolution and a $3 \times 3$ standard convolution, both activated by the GeLU function and connected in sequence, with a residual connection. The adapted Mamba features are computed as:
\begin{equation}
    f_{m^c}^{i} = \text{Res}\left([C^{3} \circ \mathcal{G}(\text{BN}(C^{1}(f_{m}^{i})))], C^{1}(f_{m}^{i})\right)
\end{equation}
where $\mathcal{G}$ denotes the GeLU activation function, $C^{k}$ represents a convolutional layer with kernel size $k$, BN denotes batch normalization, and Res$(\cdot, \cdot)$ implements the residual connection by element-wise addition. The $1 \times 1$ convolution $C^{1}$ performs channel-wise adaptation, while the $3 \times 3$ convolution $C^{3}$ captures spatial context, enabling the module to model both channel-wise relationships and spatial dependencies.

The unified features are then obtained by fusing the adapted Mamba features with the ResNet features:
\begin{equation}
\begin{split}
    f_{t}^{i} &= \mathcal{F}_{\text{MMU}}\left(f_{m}^{i}, f_{c}^{i}\right) \\
    &= \tilde{f}_{m^c}^{i} + f_{c}^{i}
\end{split}
\end{equation}
where $\tilde{f}_{m^c}^{i}$ denotes the spatially aligned version of $f_{m^c}^{i}$ to match the spatial dimensions of $f_{c}^{i}$ (if necessary), and $\mathcal{F}_{\text{MMU}}(\cdot, \cdot)$ implements the feature fusion. This unified feature $f_{t}^{i}$ combines the complementary semantic information from both heterogeneous manifolds, creating a common representation space that preserves the strengths of both architectures while enabling effective feature interaction for the subsequent student networks.

\subsection{Normal Pattern Representation Diversification}

\begin{figure*}[t]
  \centering
  \includegraphics[width=0.95\linewidth]{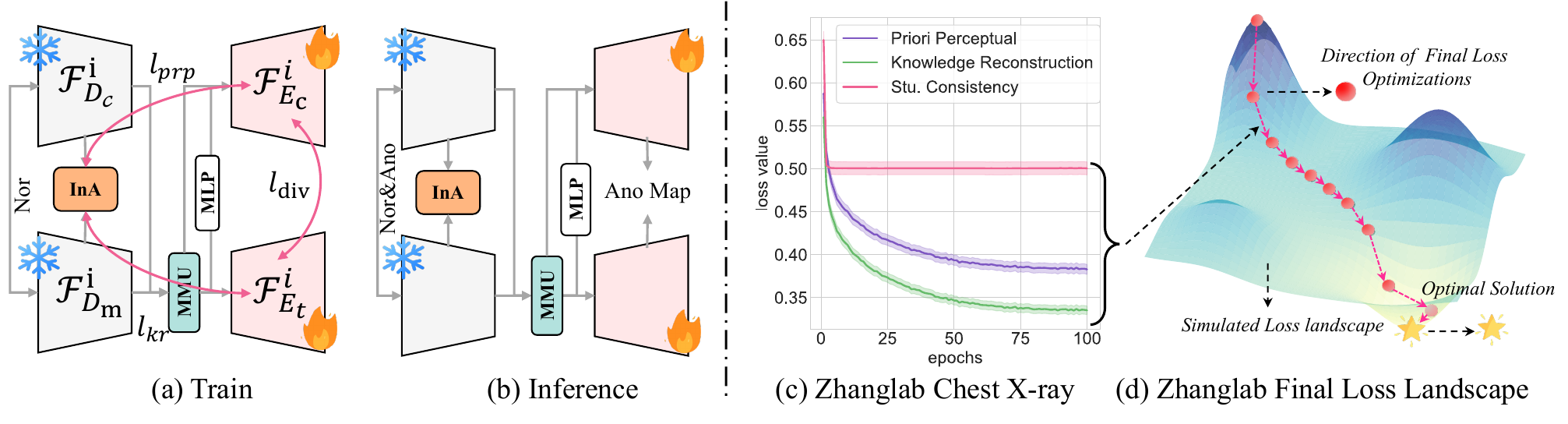}
   \caption{\textbf{Visualization of PDD pipeline and loss functions.} Figure (a) illustrates the training process of PDD, which is trained exclusively on normal samples. The training simultaneously optimizes three loss functions: $l_{prp}$, $l_{kr}$, and $l_{div}$. Figure (c) shows the loss function curves over 100 epochs on the ZhangLab Chest-Xray dataset. The combined direction of the three curves is represented in the simulated loss landscape in Figure (d), with arrows indicating the direction of loss descent. In the final 100 epochs, the training converges near the optimal solution, demonstrating that PDD effectively learns features of normal samples and can separate anomalies.}
   \label{fig:loss}
\end{figure*}

As shown in Fig.~\ref{fig:loss}-(a), to achieve diverse reconstruction of normal patterns while maintaining sensitivity to anomalies, we design a dual-student architecture with three complementary representation strategies. The unified features $f_{t}^{i}$ from the MMU module serve as the input to both student networks, enabling them to learn different aspects of normal patterns.

\textbf{Representation Strategy 1: Basic Student Network Distillation.} The first student network, denoted as Student 1, receives features from the unified manifold space $f_{t}^{i}$ and learns to reconstruct the fused features from the InA module. Formally, let $\mathcal{F}_{E_{u}}^{i}$ denote the features extracted from Student 1 at layer $i$. The distillation loss is defined as:
\begin{equation}
    \mathcal{L}_{\text{kr}} = \sum_{i=1}^{L} \|f_{b}^{i} - \mathcal{F}_{E_{u}}^{i}\|_2^2
\end{equation}
where $L$ is the number of layers, and $f_{b}^{i}$ represents the fused features from the InA module at layer $i$. This MSE-based loss ensures that Student 1 learns to capture the cross-backbone fused knowledge encoded in the InA features.

\textbf{Representation Strategy 2: Prior Knowledge-Guided Reconstruction.} The second student network, Student 2, incorporates prior knowledge from the unified manifold space through the Manifold Prior Affine module. The prior knowledge is extracted from the unified manifold features $f_{t}^{i}$ via an MLP-based affine transformation:
\begin{equation}
    z_{p}^{i} = \text{MLP}(f_{t}^{i}) = W_{p}^{i} \cdot f_{t}^{i} + b_{p}^{i}
\end{equation}
% \vspace{0.2em}
where $W_{p}^{i}$ and $b_{p}^{i}$ are learnable weight matrix and bias vector, respectively. The transformed prior knowledge $z_{p}^{i}$ is then injected into Student 2 at each layer through skip connections, enabling the network to leverage both the prior knowledge from the unified manifold and the enriched features from InA. The optimization objective combines MSE loss and cosine similarity:

\begin{equation}
    \mathcal{L}_{\text{prp}} = \sum_{i=1}^{L} \left[ \alpha \|f_{b}^{i} - \mathcal{F}_{E_{p}}^{i}\|_2^2 + \beta \left(1 - \frac{f_{b}^{i} \cdot \mathcal{F}_{E_{p}}^{i}}{\|f_{b}^{i}\|_2 \|\mathcal{F}_{E_{p}}^{i}\|_2}\right) \right]
\end{equation}

where $\mathcal{F}_{E_{p}}^{i}$ denotes the features from Student 2 at layer $i$, $\alpha$ and $\beta$ are weighting factors, and the cosine similarity term encourages angular alignment between the InA features and Student 2 features.
\begin{table*}[t]
  \centering
  \caption{Comparison with six baselines on four medical datasets. Evaluation metric: AUROC ($\%$). \textbf{Bold}: best; \underline{underline}: second best.}
%   Our model PDD and 6 medical image anomaly detection baseline models for comparison experiments, the evaluation metrics are AUROC$\uparrow$($\%$), experiments in 4 medical datasets. \textbf{Bold} numbers denote the best (SoTA) results in each row, and \underline{underlined} numbers denote the second-best results.

  \small
  \renewcommand{\arraystretch}{1.15}
  \setlength{\tabcolsep}{4pt}
  \setlength{\arrayrulewidth}{0.8pt}
  \begin{tabular}{lcccccc>{\columncolor{orange!8}}c}
  \hline
  \textbf{Method}$\rightarrow$ & \textbf{f-AnoGAN}\cite{f-AnoGAN} & \textbf{CutPaste}\cite{cutpaste} & \textbf{RD4AD}\cite{rd4ad} & \textbf{SQUID}\cite{squid} & \textbf{SIMSID}\cite{simsid} & \textbf{Skip-TS}\cite{skip-ts} & \textbf{Ours} \\
  \textbf{Dataset} $\downarrow$ & MIA'19 & CVPR'21 & CVPR'22 & CVPR'23 & TPAMI'24 & TIM'24 & \\
  \hline
  \textbf{HeadCT} & 82.6 & 73.0 & 74.3 & 75.4 & 74.9 & \underline{85.7} & \textbf{97.5} \\
  \textbf{Zhanglab} & 75.5 & 73.3 & 87.5 & 87.6 & \underline{91.1} & 79.2 & \textbf{94.0} \\
  \textbf{BrainMRI} & 77.1 & 67.0 & 80.9 & 74.7 & 81.5 & \underline{88.2} & \textbf{96.7} \\
  \textbf{CheXpert} & 65.8 & 65.5 & 71.9 & 78.1 & \textbf{79.7} & 68.7 & \underline{79.1} \\
  \hline
  \end{tabular}
  \label{tab:compare1}
  \normalsize
  \end{table*}
  
  \begin{table*}[t]
  \centering
  \caption{Comparison with SoTA methods on \textbf{Uni-Medical} dataset for multi-class anomaly localization with AU-ROC / AP / F1\_max metrics. \textbf{Bold} numbers denote the best (SoTA) results in each row, and \underline{underlined} numbers denote the second-best results.}
  \small
  \renewcommand{\arraystretch}{1.15}
  \setlength{\tabcolsep}{4pt}
  \setlength{\arrayrulewidth}{0.8pt}
  \begin{tabular}{lcccccc>{\columncolor{orange!8}}c}
  \hline
  \textbf{Method}$\rightarrow$ & \textbf{RD4AD}~\cite{rd4ad} & \textbf{UniAD}~\cite{uniad} & \textbf{SimpleNet}~\cite{liu2023simplenet} & \textbf{DeSTSeg}~\cite{destseg} & \textbf{DiAD}~\cite{diad} & \textbf{MambaAD}~\cite{mambaad} & \textbf{Ours} \\
  \textbf{Category} $\downarrow$ & CVPR'22 & NeurIPS'22 & CVPR'23 & CVPR'23 & AAAI'24 & NeurIPS'24 & \\
  \hline
  \textbf{brain} &
  82.4/94.4/91.5 &
  89.9/97.5/92.6 &
  82.3/95.6/90.9 &
  84.5/95.0/92.1 &
  \underline{93.7}/\underline{98.1}/\underline{95.0} &
  \textbf{94.2}/\textbf{98.6}/94.5 &
  90.9/96.7/\textbf{96.7} \\
  \textbf{liver} &
  55.1/46.3/64.1 &
  55.1/46.3/64.1 &
  55.8/47.6/60.9 &
  \textbf{69.2}/\textbf{60.6}/\underline{64.7} &
  59.2/\underline{55.6}/60.9 &
  \underline{63.2}/53.1/\underline{64.7} &
  59.7/50.5/\textbf{70.5} \\
  \textbf{retinal} &
  \underline{89.2}/86.7/78.5 &
  84.6/79.4/73.9 &
  88.8/87.6/78.6 &
  88.3/83.8/79.2 &
  88.3/86.6/77.7 &
  \textbf{93.6}/\underline{88.7}/\underline{86.6} &
  \textbf{93.6}/\textbf{92.7}/\textbf{88.9} \\
  \textbf{Mean} &
  75.6/75.8/78.0 &
  78.5/75.2/76.6 &
  75.6/76.9/76.8 &
  80.7/79.8/78.7 &
  80.4/\textbf{80.1}/77.8 &
  \textbf{83.7}/\textbf{80.1}/\underline{82.0} &
  \underline{81.4}/\underline{80.0}/\textbf{85.4} \\
  \hline
  \end{tabular}
  \setlength{\arrayrulewidth}{0.4pt}
  \label{tab:unimedical}
  \normalsize
  \end{table*}
\textbf{Representation Strategy 3: Student Diversity Optimization.} To prevent both student networks from collapsing into a single representation mode and to enable diverse reconstruction of normal patterns, we introduce a diversity constraint that encourages different representation strategies at different feature scales. Specifically, we apply an inverted cosine similarity constraint that encourages dissimilarity at low-dimensional feature spaces (to capture diversity) while maintaining similarity at high-dimensional feature spaces (to ensure consistency). The diversity loss is defined as:
\begin{equation}
\begin{split}
    \mathcal{L}_{\text{div}} &= \sum_{i=1}^{L_{\text{low}}} \max\left(0, \frac{\mathcal{F}_{E_{u}}^{i} \cdot \mathcal{F}_{E_{p}}^{i}}{\|\mathcal{F}_{E_{u}}^{i}\|_2 \|\mathcal{F}_{E_{p}}^{i}\|_2} - \tau_{\text{low}}\right) \\
    &\quad - \sum_{i=L_{\text{low}}+1}^{L} \min\left(0, \frac{\mathcal{F}_{E_{u}}^{i} \cdot \mathcal{F}_{E_{p}}^{i}}{\|\mathcal{F}_{E_{u}}^{i}\|_2 \|\mathcal{F}_{E_{p}}^{i}\|_2} - \tau_{\text{high}}\right)
\end{split}
\end{equation}
where $L_{\text{low}}$ denotes the number of low-dimensional layers, $\tau_{\text{low}}$ and $\tau_{\text{high}}$ are threshold parameters. The first term penalizes high cosine similarity in low-dimensional layers (encouraging diversity), while the second term penalizes low cosine similarity in high-dimensional layers (ensuring consistency). This design enables the framework to capture the diversity in medical imaging datasets, including variations in structure, density, and imaging protocols, while maintaining the ability to effectively separate anomalous patterns from normal ones.

The overall optimization objective combines the three loss functions with learnable weighting factors:
\vspace{-0.3em}
\begin{equation}
    \mathcal{L}_{\text{total}} = \lambda_{\text{kr}} \mathcal{L}_{\text{kr}} + \lambda_{\text{prp}} \mathcal{L}_{\text{prp}} + \lambda_{\text{div}} \mathcal{L}_{\text{div}}
\end{equation}
\vspace{-0.3em}
where $\lambda_{\text{kr}}$, $\lambda_{\text{prp}}$, and $\lambda_{\text{div}}$ are the weighting coefficients that balance the contributions of the knowledge distillation loss, prior knowledge loss, and diversity loss, respectively. These weights are either set as hyperparameters or learned during training to optimize the overall performance of the framework.

% \label{sec:methodology}

\section{Experiments}
% \{Experiments\}
% \label{sec:experiments}
\subsection{Dataset and Experimental Setup}

\textbf{Datasets.} We evaluate the performance of our model on multiple medical imaging datasets covering diverse anatomical regions and imaging modalities, including chest X-ray images, brain MRI images, liver CT images, and retinal OCT images. 

ZhangLab Chest X-ray~\cite{zhanglab}. This dataset contains chest X-ray images with healthy (normal) and pneumonia (anomaly) cases. Following the experimental protocol , we use 1,249 healthy images from the official training set for model training, and evaluate on a test set consisting of 234 normal and 390 abnormal images.

CheXpert~\cite{chexpert}. This large-scale chest X-ray dataset provides a comprehensive benchmark for automated chest X-ray interpretation.  We train our model on 4,499 healthy images from the official training set and test on a balanced test set containing 250 normal and 250 abnormal images.

HeadCT. This dataset consists of 200 CT images for brain hemorrhage detection. The training set contains 60 normal images, while the test set comprises 140 images including 40 normal samples and 100 abnormal brain hemorrhage samples.

BrainMRI. This dataset contains 252 MRI images for brain tumor detection. The training set includes 58 normal images, and the test set consists of 194 images with 40 normal samples and 154 abnormal samples.

Uni-Medical~\cite{Unimedical}. This multimodal dataset integrates three medical imaging benchmarks from BMAD~\cite{bmad}, encompassing brain, liver, and retinal imaging modalities. The dataset includes 7,500 training and 3,715 test images for the brain class, 1,542 training and 1,493 test images for the liver class, and 4,297 training and 1,805 test images for the retinal class.

\textbf{Metrics.} We evaluate our model at the image level using three evaluation metrics: Area Under the Receiver Operating Characteristic curve (AU-ROC), Average Precision (AP), and maximum F1 score (F1 max). Higher values indicate better performance for all these metrics.

\textbf{Implementation Details.} All input images are resized to $256 \times 256$ pixels. We train our model using the Adam optimizer with an initial learning rate of $2 \times 10^{-3}$, and the learning rate is updated using a cosine annealing schedule. All experiments are conducted on a single RTX A6000 GPU.

As illustrated in Fig.~\ref{fig:loss}-(a), during the training phase, our model is trained exclusively on normal samples. The input images are processed through the InA, MMU, and MLP modules, and the model is optimized using the loss functions described in the previous sections ($\mathcal{L}_{\text{kr}}$, $\mathcal{L}_{\text{prp}}$, and $\mathcal{L}_{\text{div}}$). During the inference phase, the trained model can perform anomaly detection on unknown samples (either normal or abnormal) by computing anomaly scores based on the reconstruction errors and feature discrepancies, as shown in Fig.~\ref{fig:loss}. 

\subsection{Comparative Experiment}

% We compare PDD against six state-of-the-art baselines: f-AnoGAN~\cite{f-AnoGAN}, CutPaste~\cite{cutpaste}, RD4AD~\cite{rd4ad}, SQUID~\cite{squid}, SIMSID~\cite{simsid}, and Skip-TS~\cite{skip-ts}. As shown in Table~\ref{tab:compare1}, PDD achieves state-of-the-art performance on three out of four datasets. On HeadCT, PDD achieves the highest AUROC of 97.5\%, outperforming Skip-TS (85.7\%) by 11.8 percentage points. On Zhanglab, PDD reaches 94.0\% AUROC, surpassing SIMSID (91.1\%) by 2.9 percentage points. On BrainMRI, PDD achieves 96.7\% AUROC, exceeding Skip-TS (88.2\%) by 8.5 percentage points. On CheXpert, PDD achieves 79.1\% AUROC, competitive with SIMSID (79.7\%).

% On the Uni-Medical dataset (Table~\ref{tab:unimedical}), PDD achieves the highest F1 max scores across all three categories: 96.7\% (brain), 70.5\% (liver), and 88.9\% (retinal). On the brain category, PDD achieves 90.9\% / 96.7\% / 96.7\% (AU-ROC / AP / F1 max), with F1 max outperforming MambaAD (94.5\%) by 2.2 percentage points. On the liver category, PDD achieves the highest F1 max of 70.5\%. On the retinal category, PDD achieves the highest AP (92.7\%) and F1 max (88.9\%), while matching the best AU-ROC (93.6\%). Overall, PDD achieves 81.4\% / 80.0\% / 85.4\% on mean performance, with F1 max outperforming MambaAD (82.0\%) by 3.4 percentage points.

%精炼后的
We compare PDD against $6$ state-of-the-art baselines: f-AnoGAN~\cite{f-AnoGAN}, CutPaste~\cite{cutpaste}, RD4AD~\cite{rd4ad}, SQUID~\cite{squid}, SIMSID~\cite{simsid}, and Skip-TS~\cite{skip-ts}. As shown in Table~\ref{tab:compare1}, PDD achieves state-of-the-art AUROC on three out of $4$ datasets—97.5\% on HeadCT, 94.0\% on Zhanglab, and 96.7\% on BrainMRI—surpassing the best baseline by 11.8, 2.9, and 8.5 percentage points, respectively. On CheXpert, PDD remains competitive with the leading method SIMSID (79.1\% vs. 79.7\%).

On the Uni-Medical dataset (Table~\ref{tab:unimedical}), PDD attains the highest F1 max across all three categories: 96.7\% (brain), 70.5\% (liver), and 88.9\% (retinal), while also achieving the best AP of 92.7\% on the retinal category. Overall, PDD surpasses the strongest competitor MambaAD by 3.4 percentage points on mean F1 max.

%单栏版本
\begin{table}[t]
    \centering
    \caption{Ablation on distillation paradigm and modular components on \textbf{ZhangLab}. Metrics: AUROC / AUPR / F1 max (\%).}
    \label{tab:ablation}
    \small
    \renewcommand{\arraystretch}{1.1}
    \setlength{\tabcolsep}{3pt}
    \resizebox{\columnwidth}{!}{%
    \begin{tabular}{l |cccc |ccc}
    \toprule
    \textbf{Model} & \makecell{InA w. MMU\\(2t1s)} & \makecell{InA w. MMU\\(2t2s)} & \makecell{RD\\(1t1s)} & MPA & \textbf{AUROC}$\uparrow$ & \textbf{AUPR}$\uparrow$ & \textbf{F1}$\uparrow$ \\
    \midrule
    M1 & \xmark & \xmark & \cmark & \xmark & 81.5 & 85.5 & 84.8 \\
    M2 & \xmark & \xmark & \cmark & \cmark & 82.0 & 86.1 & 85.7 \\
    M3 & \cmark & \xmark & \xmark & \xmark & 90.8 & 95.7 & 89.7 \\
    M4 & \cmark & \xmark & \xmark & \cmark & 92.9 & 96.6 & 90.3 \\
    Ours & \xmark & \cmark & \xmark & \cmark & \textbf{94.0} & \textbf{99.0} & \textbf{96.6} \\
    \bottomrule
    \end{tabular}%
    }
    \end{table}
% \begin{table*}[t]
% \centering
% \caption{Ablation study of the distillation paradigm and modular components on the \textbf{ZhangLab}. Evaluation metrics are image-level AUROC, AUPR, and F1 max.}
% \small
% \renewcommand{\arraystretch}{1.15}
% \setlength{\tabcolsep}{4pt}
% \begin{tabular}{l|cccc|cccc}
% \hline
% \textbf{Method} & \makecell[c]{\textbf{InA w. MMU}\\\textbf{(2t1s)}} & \makecell[c]{\textbf{InA w. MMU}\\\textbf{(2t2s)}} & \makecell[c]{\textbf{RD}\\\textbf{(1t1s)}} & \textbf{MPA} & \textbf{AUROC}$\uparrow$ & \textbf{AUPR}$\uparrow$ & \textbf{F1 max}$\uparrow$ \\
% \hline
% Model 1 & \xmark & \xmark & \cmark & \xmark & 81.5 & 85.5 & 84.8 \\
% Model 2 & \xmark & \xmark & \cmark & \cmark & 82.0 & 86.1 & 85.7 \\
% Model 3 & \cmark & \xmark & \xmark & \xmark & 90.8 & 95.7 & 89.7 \\
% Model 4 & \cmark & \xmark & \xmark & \cmark & 92.9 & 96.6 & 90.3 \\
% Final & \xmark & \cmark & \xmark & \cmark & \textbf{94.0} & \textbf{99.0} & \textbf{96.6} \\
% \hline
% \end{tabular}
% \label{tab:ablation}
% \normalsize
% \end{table*}

%单栏版本
\begin{table}[t]
    \centering
    \caption{Ablation on student consistency supervision and anomaly map computation strategies on \textbf{BrainMRI}. Metrics: AUROC / AUPR / F1 max (\%).}
    \label{tab:ablation2}
    \small
    \renewcommand{\arraystretch}{1.1}
    \setlength{\tabcolsep}{3pt}
    \resizebox{\columnwidth}{!}{%
    \begin{tabular}{l |ccc |ccc}
    \toprule
    \textbf{Model} & $L_{div}$ & \makecell{$cos(t1,s1)$\\$+cos(t2,s2)$} & $cos(s1,s2)$ & \textbf{AUROC}$\uparrow$ & \textbf{AUPR}$\uparrow$ & \textbf{F1}$\uparrow$ \\
    \midrule
    M1 & \cmark & \xmark & \cmark & 32.50 & 81.23 & 93.02 \\
    M2 & \xmark & \xmark & \cmark & 70.46 & 97.26 & 95.65 \\
    M3 & \xmark & \cmark & \xmark & 93.41 & 98.81 & 95.93 \\
    Ours & \cmark & \cmark & \xmark & \textbf{96.67} & \textbf{99.51} & \textbf{97.56} \\
    \bottomrule
    \end{tabular}%
    }
    \end{table}

% \begin{table*}[htbp]
% \centering
% \caption{Ablation on student consistency supervision and anomaly map computation strategies on \textbf{BrainMRI}. Evaluation metrics are image-level AUROC, AUPR, and F1 max.}
% \small
% \renewcommand{\arraystretch}{1.15}
% \setlength{\tabcolsep}{4pt}
% \begin{tabular}{l|ccc|ccc}
% \hline
% \textbf{Method} & \textbf{$L_{div}$} & \textbf{$cos(t1,s1)+cos(t2,s2)$} & \textbf{$cos(s1,s2)$} & \textbf{AUROC}$\uparrow$ & \textbf{AUPR}$\uparrow$ & \textbf{F1 max}$\uparrow$ \\
% \hline
% Model 1 & \cmark & \xmark & \cmark & 32.50 & 81.23 & 93.02 \\
% Model 2 & \xmark & \xmark & \cmark & 70.46 & 97.26 & 95.65 \\
% Model 3 & \xmark & \cmark & \xmark & 93.41 & 98.81 & 95.93 \\
% Final & \cmark & \cmark & \xmark & \textbf{96.67} & \textbf{99.51} & \textbf{97.56} \\
% \hline
% \end{tabular}
% \label{tab:ablation2}
% \normalsize
% \end{table*}

% \caption{Ablation study on unsupervised anomaly localization: training-time student consistency supervision ($L_{div}$) and inference-time anomaly map computation strategies ($cos(t1,s1)+cos(t2,s2)$ and $cos(s1,s2)$) on the \textbf{BrainMRI} dataset. Evaluation metrics are image-level AUROC, AUPR, and F1 max.}

%美观
\begin{table}[t]
    \centering
    \caption{Ablation on $\tau_{\text{low}}$ and $\tau_{\text{high}}$ across datasets. Metric: AUROC (\%).}
    \label{tab:tau_ablation}
    \small
    \renewcommand{\arraystretch}{1.1}
    \setlength{\tabcolsep}{3pt}
    \resizebox{\columnwidth}{!}{%
    \begin{tabular}{l| ccc |ccc |c}
    \toprule
     & \multicolumn{3}{c}{\textbf{ZhangLab}} & \multicolumn{3}{c}{\textbf{HeadCT}} & \textbf{BrainMRI} \\
    \cmidrule(lr){2-4} \cmidrule(lr){5-7} \cmidrule(lr){8-8}
    $\tau_{\text{high}}$ & 0.40 & 0.30 & 0.75 & 0.30 & 0.30 & 0.75 & 0.75 \\
    $\tau_{\text{low}}$  & 0.65 & 0.75 & 0.40 & 0.65 & 0.75 & 0.40 & 0.40 \\
    \midrule
    AUROC & 94.52 & \textbf{96.18} & 95.92 & 96.92 & \textbf{97.31} & 96.15 & \textbf{93.30} \\
    \bottomrule
    \end{tabular}%
    }
    \end{table}
% \begin{table}[htbp]
%     \centering
%     \footnotesize
%     \setlength{\tabcolsep}{3pt}
%     \renewcommand{\arraystretch}{0.85}
%     \caption{Ablation study on $\tau_{\text{low}}$ and $\tau_{\text{high}}$ across datasets.}
%     \label{tab:tau_ablation}
%     \begin{tabular}{@{}lccccccc@{}}
%     \toprule
%      & \multicolumn{3}{c}{ZhangLab} & \multicolumn{3}{c}{HeadCT} & BrainMRI \\
%     \cmidrule(lr){2-4} \cmidrule(lr){5-7} \cmidrule(lr){8-8}
%     $\tau_{\text{high}}$ & 0.40 & 0.30 & 0.75 & 0.30 & 0.30 & 0.75 & 0.75 \\
%     $\tau_{\text{low}}$ & 0.65 & 0.75 & 0.40 & 0.65 & 0.75 & 0.40 & 0.40 \\
%     AUROC & 94.52 & \textbf{96.18} & 95.92 & 96.92 & \textbf{97.31} & 96.15 & \textbf{93.30} \\
%     \bottomrule
%     \end{tabular}
%     \end{table}

\begin{figure}[htb]
    \centering
    \includegraphics[width=0.95\linewidth]{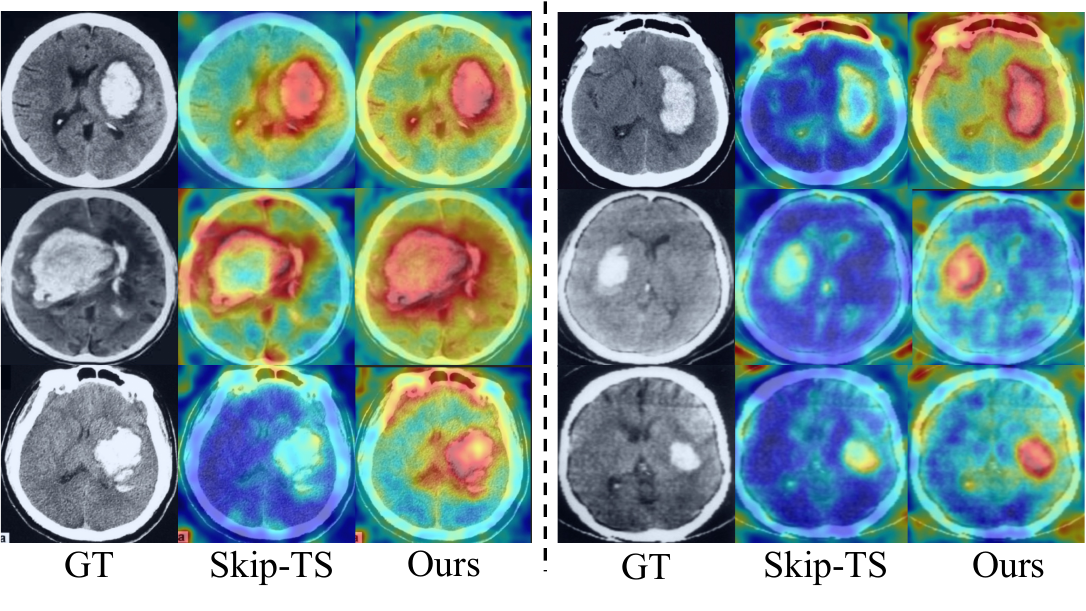}
     \caption{Anomaly localization comparison between PDD and Skip-TS on the HeadCT dataset.}
     \label{fig:headct}
\end{figure}

\begin{figure}[htb]
    \centering
    \includegraphics[width=0.95\linewidth]{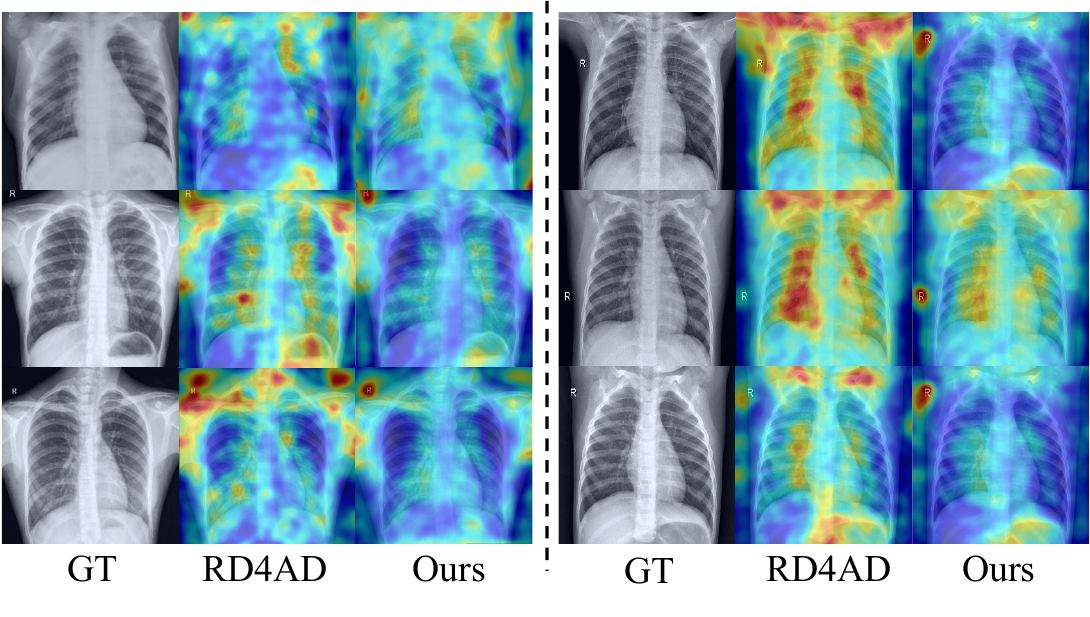}
     \caption{Anomaly localization comparison between PDD and RD4AD on the ZhangLab dataset. PDD produces significantly fewer false positives on normal samples, demonstrating stronger specificity.}
     \label{fig:fake}
\end{figure}

\subsection{Ablation Study}

To validate the effectiveness of each component in our proposed PDD framework, we conduct comprehensive ablation studies on the ZhangLab Chest-Xray and BrainMRI datasets. Table~\ref{tab:ablation} presents the ablation results for different distillation paradigms and modular components.

\textbf{Ablation on Distillation Paradigms and Modular Components.} 
We evaluate different architectural configurations in Table~\ref{tab:ablation}, where $(2t1s)$ and $(2t2s)$ denote two teachers with one/two students, respectively. Starting from a vanilla RD(1t1s) baseline (Model 1: 81.5\% / 85.5\% / 84.8\%), introducing the dual-teacher architecture with InA and MMU (Model 3) yields a substantial gain of 9.3 percentage points in AUROC, validating the benefit of complementary representations. Adding MPA (Model 4) further improves AUROC to 92.9\%, and upgrading to a dual-student design (our final model) achieves the best performance of 94.0\% / 99.0\% / 96.6\%. These results confirm that: (1) dual-teacher architecture with InA and MMU effectively leverages complementary representations; (2) dual-student design significantly enhances diverse normal pattern learning; and (3) MPA provides useful prior knowledge for further improvement.

\textbf{Ablation on Anomaly Localization.} We investigate the effects of training-time student consistency supervision and inference-time anomaly map computation on BrainMRI, as shown in Table~\ref{tab:ablation2}. Three components are examined: diversity loss \textit{$L_{div}$} between two students during training, teacher-student cosine similarity \textit{$cos(t1,s1)+cos(t2,s2)$} for inference, and student-student cosine similarity \textit{$cos(s1,s2)$} for inference. Using only \textit{$cos(s1,s2)$} with \textit{$L_{div}$} (Model 1) leads to poor AUROC of 32.50\%, while removing \textit{$L_{div}$} (Model 2) raises it to 70.46\%, suggesting that \textit{$L_{div}$} forces divergent student representations that harm student-student based localization. Switching to teacher-student alignment (Model 3) significantly boosts AUROC to 93.41\%. Our final model combines \textit{$L_{div}$} with teacher-student alignment, achieving the best performance of 96.67\% / 99.51\% / 97.56\%. This demonstrates that: (1) teacher-student alignment is essential for effective anomaly localization; (2) \textit{$L_{div}$} prevents student collapse into identical representations; and (3) the combination of diversity supervision and teacher-student alignment achieves optimal results.

\textbf{Hyper-parameter Sensitivity.} We further conduct hyper-parameter ablations on the diversity thresholds $(\tau_{\text{low}}, \tau_{\text{high}})$ and the loss weights $(\lambda_{\text{kr}}, \lambda_{\text{prp}}, \lambda_{\text{div}})$. As shown in Table~\ref{tab:tau_ablation}, PDD achieves consistently strong AUROC across datasets under different $(\tau_{\text{low}}, \tau_{\text{high}})$ combinations, indicating that the performance is not overly sensitive to these thresholds and the method is robust to moderate parameter changes. We set $\tau_{\text{low}}=0.30$ and $\tau_{\text{high}}=0.75$ in all experiments. For the overall objective, we use $\lambda_{\text{kr}}=\lambda_{\text{prp}}=0.02$ and $\lambda_{\text{div}}=0.5$, which provides a good balance between distillation, prior-guided reconstruction, and diversity regularization.

\subsection{Anomaly Localization}
Our anomaly localization is achieved during inference by aggregating the cosine similarity between each teacher-student pair, formulated as $\cos(t_1,s_1)+\cos(t_2,s_2)$.

% Figure~\ref{fig:headct} presents qualitative anomaly localization results on the HeadCT dataset, comparing our proposed PDD method with Skip-TS~\cite{skip-ts}. The visualization consists of six representative test cases arranged in three columns: ground truth annotations (left), Skip-TS predictions (middle), and PDD predictions (right). As demonstrated in the figure, our method exhibits superior performance particularly in challenging scenarios characterized by irregular lesion boundaries and subtle pathological patterns. In cases where anomalies are less conspicuous or exhibit irregular spatial distributions, PDD demonstrates more accurate and precise localization compared to Skip-TS, effectively capturing nuanced pathological regions that may be overlooked by the baseline method.
%精炼版本
Figure~\ref{fig:headct} presents qualitative anomaly localization results on HeadCT, comparing PDD with Skip-TS~\cite{skip-ts}. Each column shows ground truth (left), Skip-TS predictions (middle), and PDD predictions (right). Our method achieves more accurate localization, particularly for lesions with irregular boundaries and subtle pathological patterns that are often overlooked by Skip-TS.
% Please refer to the supplementary material for detailed anomaly localization results and visualizations.

Figure~\ref{fig:fake} compares anomaly localization between PDD and RD4AD on the ZhangLab dataset. As shown, RD4AD tends to produce spurious activations on normal samples, while PDD maintains cleaner predictions with significantly fewer false positives, indicating stronger specificity in distinguishing normal regions from anomalies.
\section{Conclusion}

% This paper addresses the unique challenges of medical image anomaly detection by proposing \textbf{PDD}, a manifold-prior diverse distillation framework. Through systematic Grad-CAM analysis, we reveal the fundamental limitations of discriminative activation maps on medical data, motivating the need for manifold-level modeling. Our framework unifies complementary priors from VMamba-Tiny and wide-ResNet50 encoders into a shared high-dimensional manifold through the \textbf{MMU} module, while the \textbf{InA} module enriches intermediate feature representations. The dual-student architecture, combined with \textbf{MPA}-based prior knowledge injection and diversity regularization, enables diverse reconstruction of normal patterns while maintaining sensitivity to subtle anomalies. Extensive experiments demonstrate that \textbf{PDD} significantly outperforms existing methods, achieving improvements of up to 11.8\%, 5.1\%, and 2.9\% in AUROC on HeadCT, BrainMRI, and ZhangLab datasets, respectively, and 3.4\% in F1 max on the Uni-Medical dataset, establishing new state-of-the-art performance in medical image anomaly detection.

We propose \textbf{PDD}, a manifold-prior diverse distillation framework for medical image anomaly detection. By unifying complementary priors from VMamba-Tiny and wide-ResNet50 through the \textbf{MMU} and \textbf{InA} modules, and leveraging a dual-student architecture with \textbf{MPA}-based prior injection and diversity regularization, PDD enables diverse reconstruction of normal patterns while preserving sensitivity to subtle anomalies. Extensive experiments across multiple benchmarks demonstrate that PDD achieves state-of-the-art performance, validating the effectiveness of manifold-level prior modeling for medical anomaly detection.

Despite its strong performance, PDD still has limitations. In particular, the model tends to produce false positives on non-pathological artifacts commonly present in medical images, such as imaging device markers and implanted metallic objects. These artifacts exhibit visual patterns that deviate from normal tissue but are clinically irrelevant, posing challenges for anomaly detection methods that rely on appearance-level deviation. Addressing this limitation, for instance through artifact-aware prior modeling or clinical context integration, remains an important direction for future work.
{
    \small
    \bibliographystyle{ieeenat_fullname}
    \bibliography{main}
}


\end{document}